\documentclass[runningheads]{llncs}
\usepackage{graphicx,wrapfig,subcaption}

\usepackage{tikz}
\usepackage{comment}
\usepackage{amsmath,amssymb} 
\usepackage{color}

\usepackage[accsupp]{axessibility}  
\usepackage{amsmath,amssymb,amsfonts}

\newcommand{\X}{\mathbf{X}}
\newcommand{\T}{\mathbf{T}}
\newcommand{\Xflip}{\mathbf{X_{f}}}
\newcommand{\Y}{\mathbf{Y}}
\newcommand{\Yflip}{\mathbf{Y_{f}}}

\begin{document}
\pagestyle{headings}
\mainmatter
\def\ECCVSubNumber{2750}  

\title{Joint Symmetry Detection and Shape Matching for Non-Rigid Point Clouds} 

%
\author{Abhishek Sharma \and Maks Ovsjanikov}
\authorrunning{A. Sharma \and M. Ovsjanikov}
%
\institute{LIX, Ecole Polytechnique, IP Paris 
 }
\maketitle
\begin{abstract}
   Despite the success of deep functional maps in non-rigid 3D shape matching, there exists no learning framework that models both self-symmetry and shape matching simultaneously. This is despite the fact that errors due to symmetry mismatch are a major challenge in non-rigid shape matching. In this paper, we propose a novel framework that simultaneously learns both self symmetry as well as a pairwise map between a pair of shapes. Our key idea is to couple a self symmetry map and a pairwise map through a regularization term that provides a joint constraint on both of them, thereby, leading to more accurate maps. We validate our method on several benchmarks where it outperforms many competitive baselines on both tasks.
\end{abstract}
\section{Introduction}
\label{sec:intro}
Shape correspondence is a fundamental problem in computer vision, computer graphics and related fields \cite{Thomas21}, since it facilitates  many high level tasks such as texture or deformation transfer. Although shape correspondence has been studied from many viewpoints \cite{van2011survey}, we focus here on a functional map-based approaches \cite{ovsjanikov2012functional} as this framework is quite general, scalable and thus, has been extended to various other applications such as pose estimation \cite{neverova20}, matrix completion \cite{sharmamatrix} and graph matching \cite{FRGM20}.

While recent learning-based deep functional map approaches have made impressive gains in non-rigid isometric full shape matching \cite{litany17,roufosse2019unsupervised,halimi2018self,sharma20}, symmetry detection \cite{Nagar18,jing_sig20} has received little attention in the learning paradigm. This is despite the fact that the two problems are inherently linked and symmetry disambiguation remains a challenge in shape matching pipelines (see Figure 1). Sharma $\&$ Ovsjanikov \cite{sharma20} proposes to align the shapes rigidly and claims empirically that manual rigid alignment resolves some symmetry problems arising in shape matching. In this work, we go a step further and provide an alternative based on a principled approach that simultaneously models both self symmetry and shape matching and thereby, considers the intrinsic self-symmetry during training. Moreover, while there have been some attempts to learn shape matching in a noisy setup \cite{litany17,ric_linear20}, we are not aware of any learning setup that investigates the effect of noise for symmetry detection.

Our key contribution is a novel commutative regularization that couples the self-symmetry map with a pairwise map and thus, enables knowledge transfer between the two maps during training. This significantly improves generalization and robustness to sampling resolution as well as the size of embedding. Our method obtains competitive results on multiple shape matching benchmarks such as FAUST remesh \cite{ren2018continuous} and partial SHREC'16 \cite{cosmo2016matching} when compared to recent learning-based methods while being very robust to noisy set up. We also evaluate our method on symmetry detection on various benchmarks where it shows resilience to noise where other methods based on the Laplacian Beltrami operator fail. 
\begin{figure} 
\minipage{0.36\textwidth}
  \includegraphics[width=\linewidth]{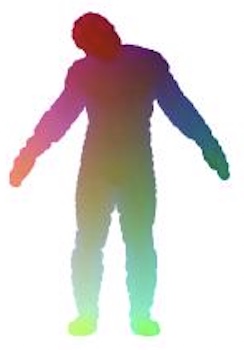}
  \subcaption{ Source shape}
\endminipage
\minipage{0.25\textwidth}
  \includegraphics[width=\linewidth]{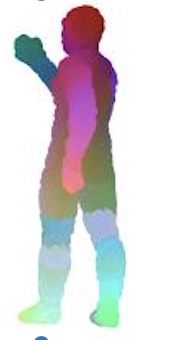}
  \subcaption{\cite{sharma20}}
\endminipage
\minipage{0.25\textwidth}%
 \includegraphics[width=\linewidth]{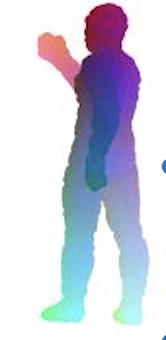}
  \subcaption{Ours.}
\endminipage
\caption{On the left, we show a source shape that is to be matched with a target shape. In the middle, we show the color coded map on target shape using \cite{sharma20} whereas on the right, we show our color coded map. Notice that \cite{sharma20} completely mismatches left and right hand, our method does not.}
\label{fig:intro}
\end{figure}
To summarize, our contributions are as follows: \begin{itemize}
    
    \item To the best of our knowledge, we propose a first method that simultaneously learns symmetry detection and shape matching  for non-rigid point clouds.
    
    \item We propose a novel regularization that constrains the symmetry map and pairwise map which is of independent interest for future work in this direction.
\end{itemize}

\section{Related Work}
\label{sec:related}

\paragraph{Functional Maps} Computing point-to-point maps between two 3D discrete surfaces is a very well-studied problem. We refer to a recent survey \cite{sahilliouglu2019recent}  for an in-depth discussion. Our method is closely related to the functional map pipeline, introduced in \cite{ovsjanikov2012functional} and then significantly extended in follow-up works (see, e.g.,\cite{ovsjanikov2017computing}). The key idea of this framework is to encode correspondences as small matrices, by using a reduced functional basis, thus greatly simplifying many resulting  optimization problems. The functional map pipeline has been further improved in accuracy, efficiency and robustness by many recent works including ~\cite{kovnatsky2013coupled,huang2014functional,burghard2017embedding,rodola2017partial,commutativity,ren2018continuous,CyclicFM20}. There also exist other works \cite{wei2016dense,MasBosBroVan16,monti2017} that treat shape correspondence as a dense labeling problem but they typically require a lot of data as the label space is very large. 
\paragraph{Learning from raw 3D shape}
Although early approaches in functional maps literature used hand-crafted features \cite{ovsjanikov2017computing}, more recent methods aim to \emph{learn} either the optimal transformations of hand crafted descriptors \cite{litany17,roufosse2019unsupervised} or even  features directly from 3D geometry itself \cite{sharma20}. Initial efforts in this direction used classical optimisation techniques \cite{corman2014supervised}. In contrast, Deep Functional Maps \cite{litany2017deep} proposed a deep learning architecture called FMNet to optimize a non-linear transformation of SHOT descriptors \cite{shot}, that was further extended to unsupervised setting \cite{roufosse2019unsupervised,halimi2018self,deepshells}. To alleviate the sensitivity of the SHOT descriptor to changes in mesh structure, recent works including \cite{groueix20183d,sharma20} learn shape matching directly from the \emph{raw 3D data} without relying on pre-defined descriptors, thus leading to improvements in both robustness and accuracy. However, all these works are aimed at \textit{full} (complete) shape correspondence and do not handle partial shape matching effectively. Our work is also related to a recent work \cite{ric_linear20} that proposes to replace the Laplace-Beltrami basis with learned embeddings. However, unlike \cite{ric_linear20}, we do not impose linearly invariant constraint between the learned embeddings.
\paragraph{Symmetry for Non Rigid Shape Matching} Matching shapes with intrinsic symmetries involves dealing with symmetric ambiguity problem which has been very well studied and explored in axiomatic methods~\cite{Raviv10,sym_Lipman10,mitrastar12,maks_quotient13,Nagar18,jing_sig20}. More recently, \cite{sym-img,clara20} proposes an end to end method to learn extrinsic 3D symmetries from a RGB-D image. However, none of the existing learning based non-rigid shape matching method models or learn symmetry explicitly as a regularizer for shape matching. 
\paragraph{Joint Learning of similar tasks} Computer vision literature is full of problems that are inherently linked~\cite{ubernet,sharma18,neuromorph} and thus, should be learned simultaneously. In 3D shape analysis, Neuromorph~\cite{neuromorph} simultaneously learns shape correspondence and interpolation. Our work also follows a similar direction as we aim to learn shape matching and symmetry detection simultaneously. Our work is most similar in spirit to \cite{sharma18} that couples image segmentation and detection via linear constraints and thus, induces information transfer/sharing between the segmentation map and detection map via these constraints. In our formulation, we enable this information transfer during training via a commutative loss that couples the self-symmetry and pairwise map.

 \paragraph{} The rest of this paper is structured as follows: In the next section, we first propose our method to learn canonical embedding for joint shape matching and symmetry detection and introduce our novel regularization term that constrains self-symmetry and pairwise map. We then consider the unsupervised setting in which symmetry supervision is not provided. Lastly, we validate our framework on three benchmark datasets by comparing it to various state-of-the-art methods and providing ablation studies.
 
\section{Joint Shape Matching and Symmetry Detection}
\label{sec:method}
 Due to the instability of Laplace-Beltrami operator, LBO, on partial 3D shapes~\cite{kirgo2020wavelet} and noise~\cite{ric_linear20}, our main goal is to avoid using its eigenfunctions and instead we aim to \emph{learn} an embedding that can replace the spectral embedding given by the LBO. This section details how to learn such an embedding while working in the symmetric space.

\paragraph{Input Shape Representation} In contrast to several recent works \cite{halimi2018self,sharma20} that assume to be given a mesh representation of 3D shapes in terms of LBO operator, we do not impose such a constraint and work directly with the point cloud representation. We denote a map between a pair of shapes $\X$ and $\Y$ by $T_{\X\Y}:\X \rightarrow \Y$ so that $T_{\X\Y}(x_i) = y_j$, $\forall i \in \{1, \ldots , n_{\X}\}$ and some $j  \in \{1, \ldots , n_{\Y}\}$.  This map can be represented by a matrix $\Pi_{\X\Y} \in \mathbb{R}^{n_{\X}\times n_{\Y}}$ such that $\Pi_{\X\Y}(i,j) = 1$ if  $T_{\X\Y}(x_i) = y_j$ and $0$ otherwise. We use $P_{\X}$ to denote the 3D coordinates of $\X$. 

\subsection{Supervised Loss functions}
In the supervised setting, we assume to be given a set of pairs of shapes $\X,\Y$ for which ground truth correspondences $\T^{gt}_{\X\Y}$ as well as the ground truth self-symmetry map  $\T^{sym}$ are known. Our main goal in the supervised setting is to model intrinsic symmetry using ground truth  symmetry map $\T^{sym}$ which will ultimately lead to better shape matching. Our network takes input as $P_{\X}$, 3D coordinates of point clouds, computes an embedding  $\Phi_{\X} \in \mathbb{R}^{n_{\X}\times k}$ for each shape based on a PointNet~\cite{qi2017pointnet} feature extractor  that embeds the shapes into some fixed $k$ dimensional space. The parameters of this feature extractor are learned by optimizing the sum of  two loss functions during training as detailed next. 

\paragraph{Cosine Similarity} Our loss functions are based on a soft-correspondence matrix, also used in \cite{litany2017deep} and \cite{ric_linear20}. The \textit{soft} correspondence matrix $S_{\X\Y}$  is a soft version of the binary correspondence matrix $\Pi_{\X\Y}$.
We compare the rows of $\Phi_{\X}$ to those of $\Phi_{\Y}$ to obtain the \textit{soft} correspondence matrix $S_{\X\Y}$ that approximates the pairwise map in a differentiable way as follows:
\begin{equation}
    \begin{aligned}
        (S_{\X\Y})_{ij} = \frac{e^{\Phi_{\X}^{{i}^T} \Phi_{\Y}^{j}/\tau}}{\sum_{j} e^{\Phi_{\X}^{{j}^T} \Phi_{\Y}^{j}/\tau}} 
    \end{aligned}
    \label{eq:softmax}
\end{equation}
%
where $\Phi_{\X}^{{i}^T} \Phi_{\Y}^{j}$ measures the similarity between any two pointwise embeddings and is defined as their inner product where the scalar $\tau$ is set to $.3$.  

\noindent\textbf{Nearest Neighbour Loss}
Our Nearest Neighbour loss links the embeddings of the two shapes and is designed to preserve the given ground truth mapping. Specifically, we first compute the soft correspondence matrix $S_{\X\Y}$ between a pair of shapes, by comparing the rows of $\Phi_{\X}$ to those of $\Phi_{\Y}$ in a differentiable way as done in Eq. \eqref{eq:softmax}. We then evaluate the computed soft map, again, by evaluating how well it transfers the coordinate functions, compared to the given ground truth mapping.
\begin{equation}
L(\Phi_{\X},\Phi_{\Y})_{NN.} = \sum \|  S_{\X\Y} P_{\Y} - \T^{gt}_{\X\Y} P_{\Y} \|_2^2.
\label{eq:loss_euc}
\end{equation}
Note that unlike the linearly invariant loss imposed in \cite{ric_linear20}, this loss is based on comparing $\Phi_{\X}$ and $\Phi_{\Y}$ directly, without computing any linear transformations. This significantly simplifies the learning process and in particular, reduces the computation of the correspondence at test time to a simple nearest-neighbor search. Despite this, as we show below, due to our strong regularization, our approach achieves superior results compared to the method of \cite{ric_linear20}, based on computing an optimal linear transformation at test time.

\noindent\textbf{Symmetry Commutativity Loss} Our next loss aims to link the symmetry map computed for each shape and the correspondence across the two shapes. We achieve this by using the algebraic properties of the functional representation, and especially using the fact that map composition can simply be expressed as matrix multiplication.

Specifically, given a self-symmetry pointwise groundtruth maps on shape $\X$ and shape $\Y$, we aim to promote the \textit{consistency} between the computed correspondence and the symmetries on each shape. We do this by imposing the following commutativity loss during training:
\begin{equation}
     L(\Phi_{\X},\Phi_{\Y})^{sup}_{comm.} =  \|\T^{sym}_{\X} S_{\X\Y} - S_{\X\Y} \T^{sym}_{\Y} \|_2
     \label{eq:comm}
 \end{equation}

Intuitively, this loss considers the difference between mapping from $\X$ to $\Y$ and applying the symmetry map on $\Y$, as opposed to applying the symmetry on $\X$ and then mapping from $\X$ to $\Y$. Note that this is similar to the commonly used \textit{Laplacian} commutativity in the functional map literature \cite{ovsjanikov2012functional}. However, rather than promoting isometries, our loss enforces that the computed map respects the self-symmetry structure of each shape, which holds regardless of the deformation class, and is not limited to isometries.

\paragraph{Overall Loss} We combine the two embedding losses defined in \eqref{eq:loss_euc} and \eqref{eq:comm} and write the overall loss as follows:
\begin{equation}
     L_{sup.} =  L_{NN.} + \gamma *L^{sup}_{comm.}   
     \label{eq:tot}
 \end{equation}
\subsection{Unsupervised Setting}

\begin{wrapfigure}{r}{0.5\textwidth}
\centering
\minipage{0.24\textwidth}
  \includegraphics[width=\linewidth]{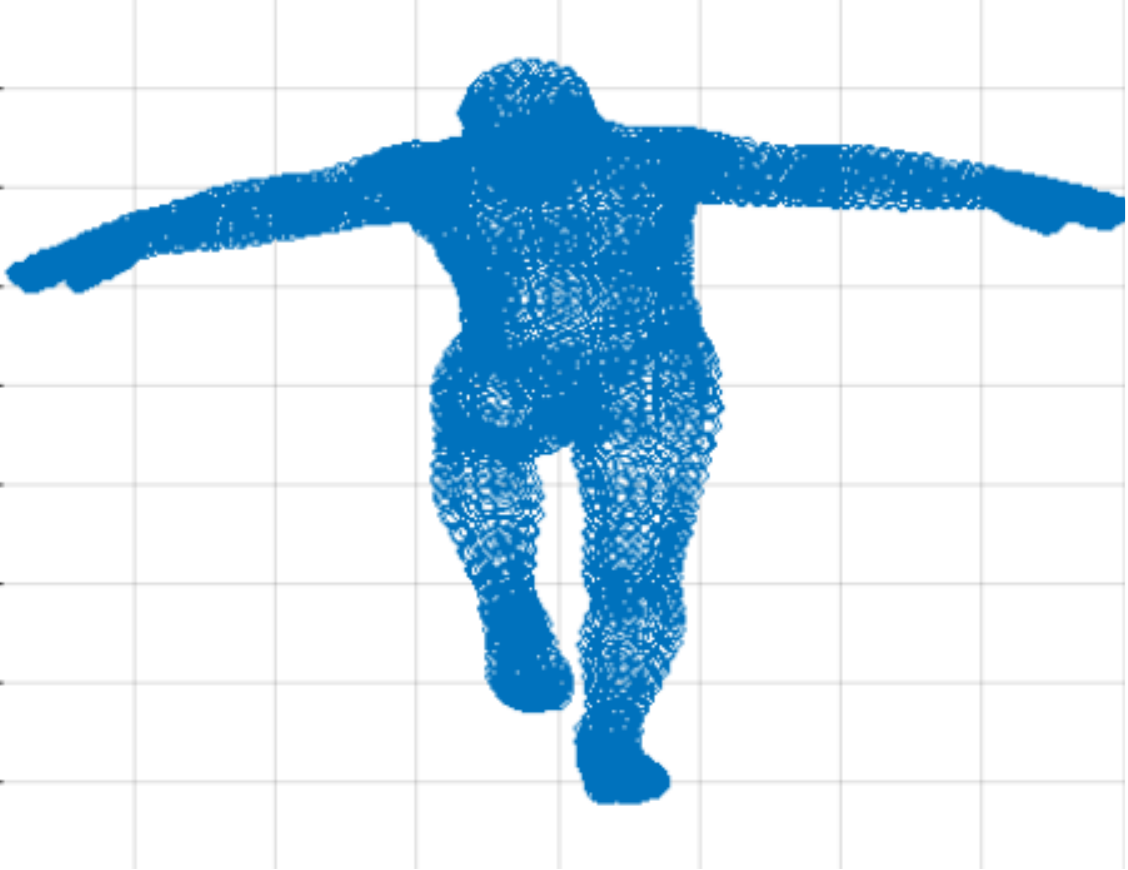}
\endminipage%
\minipage{0.24\textwidth}
  \includegraphics[width=\linewidth]{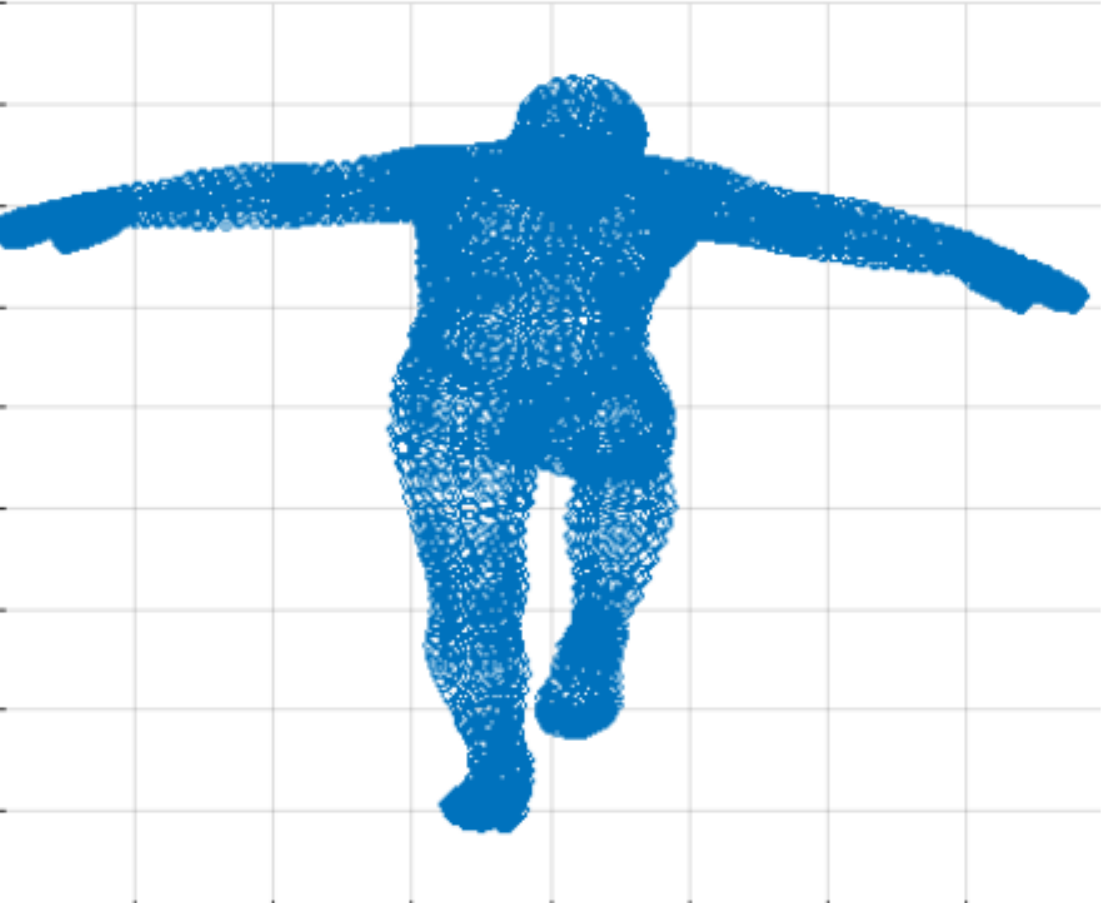}
\endminipage
\caption{On the left, we show a source shape and on the right, we show 
the flipped version.}
\label{fig:flip}
\end{wrapfigure}

Our network takes a shape $\X$ as input. We then perform a reflection (flip) of each shape along X-axis resulting in a shape denoted as $\Xflip$. We also experimented with other axis but chose a flip along X axis as most of the datasets by default have a symmetry bias along this axis and thus, best performance. The original and flipped shapes are then forwarded to a Siamese architecture, based on a PointNet~\cite{qi2017pointnet} feature extractor,  that embeds these two shapes into some fixed $k$ dimensional space. We illustrate in Figure \ref{fig:flip} one such flip. 
The intuition behind such extrinsic flip is to let the network learn two different embeddings for the same shape from which a symmetry map can be computed when no symmetry ground truth is given. Let $\Phi_{\X}$ and $\Phi_{\Xflip}$ denote the matrices, whose rows can be interpreted as embeddings of the points of $\X$ and $\Xflip$.

\textbf{Self-Symmetry Map} We compare the rows of $\Phi_{\X}$ to those of $\Phi_{\Xflip}$ to obtain the \textit{soft} correspondence matrix $S_{\X\Xflip}$ that approximates the self-symmetry map in a differentiable way as follows:
\begin{equation}
    \begin{aligned}
        (S_{\X\Xflip})_{ij} = \frac{e^{\Phi_{\X}^{{i}^T} \Phi_{\Xflip}^{j}/\tau}}{\sum_{j} e^{\Phi_{\X}^{{j}^T} \Phi_{\Xflip}^{j}/\tau}} 
    \end{aligned}
    \label{eq:softmax_sym}
\end{equation}

\textbf{Unsupervised Symmetry Commutativity} 
We enforce a \textit{consistency} between the computed correspondence and the estimated symmetry map in an unsupervised way using commutativity loss as follows:
\begin{equation}
     L(\Phi_{\X},\Phi_{\Y})^{uns}_{comm.} =  \|S_{\Xflip\X} S_{\X\Y} - S_{\X\Y} S_{\Y\Yflip} \|_2
     \label{eq:comm_uns}
 \end{equation}
 
 \paragraph{Overall Loss} We combine the two embedding losses defined in \eqref{eq:loss_euc} with that of commutativity loss defined in \eqref{eq:comm_uns} and define the training loss as follows:
 \begin{equation}
     L_{unsup.} =  L_{NN.} + \gamma *L^{uns}_{comm.}  
     \label{eq:tot_uns}
 \end{equation}

The scalar $\gamma$ allows us to weigh the symmetry information differently in a supervised setting where we assume to be given a self-symmetry map and in an unsupervised setting where we work without a symmetry map. Naturally, we set it higher for the supervised case where enforcing symmetry structure makes more sense than unsupervised case where symmetry is induced by a pairwise matching NN loss and transferred by commutativity loss. We set $\gamma$ to $1$ for supervised setting and $.2$ for unsupervised setting.

\paragraph{Test Phase} At test time, once the network is trained, we simply compute the embedding $\Phi_{\X}$ and $\Phi_{\Y}$ and do a nearest neighbour search between them to find correspondence between the two shapes. Similarly, to estimate a self-symmetry map, we compute the embedding $\Phi_{\X}$ and $\Phi_{\Xflip}$ and do a nearest neighbour search between them.

\paragraph{Implementation Details}
We implemented our method in Pytorch \cite{pytorch}. All experiments are run on a Nvidia RTX $2080$ graphics processing card and require $16$ GB of GPU memory. We learn a $k=20$ dimensional embedding (basis) for each point cloud. Following \cite{sharma20,ric_linear20}, our feature extractor is also based on the architecture of PointNet. We use a batch size of $8$ and learning rate of $1e-4$ and optimize our objective with Adam optimizer in Pytorch \cite{pytorch}. 
During training, we randomly sample $3000$ points from the point cloud and obtain an embedding of $20$ dimensions. Our results are not sensitive to small changes in these two parameters. We experimented with an embedding size of $20,40,60$ and obtained an average geodesic error in the range $33-36$ on FAUST-R. Similarly, in addition to the $3000$ point cloud resolution during training, we  also tried a point cloud resolution in the range $2k-4k$ and found almost negligible drop in performance. This can be explained by Pointnet resilience to change in point cloud density.

\section{Results}
\label{sec:results}
This section is divided into three subsections. First subsection \ref{subsec:results_full} shows the experimental comparison of our approach with state-of-the art methods for shape matching and tests our method on a wide spectrum of datasets: the re-meshed versions \cite{ren2018continuous} of FAUST dataset \cite{bogo2014} and  SHREC’16 Partial Correspondence dataset \cite{cosmo2016matching}. These experiments validate the promising direction of our embedding based method as it obtains competitive performance on these two benchmarks and especially outperforms LBO based methods on benchmarks with noise. The next subsection \ref{subsec:ablation} ablates the overall performance and experimentally validates our claim that shape matching with canonical embedding with appropriate regularization outperforms the linearly invariant embeddings proposed in Marin et al.  \cite{ric_linear20}. We demonstrate this with both symmetry supervision as well as without symmetry supervision. Lastly, Section \ref{subsec:symmetry} shows the effectiveness of our method on the symmetry detection task in the presence of noise. We evaluate all results by reporting the per-point-average geodesic distance between the ground truth map and the computed map. All results are multiplied by 100 for the sake of readability.  We conclude with an illustration showing a failure case of our method. We provide more qualitative comparison results in supplement.

\subsection{Shape Matching}\label{subsec:results_full} We present our results on a full shape matching benchmark dataset FAUST remesh\cite{bogo2014,ren2018continuous}, denoted in future subsections as FAUST-R. We also use its two other versions used previously: the Faust aligned dataset used in \cite{sharma20}, denoted as FAUST-A and noisy Faust version \cite{ric_linear20} denoted as FAUST-N. All these datasets contain $100$ shapes of $10$ different subjects in different poses where each point cloud contains roughly $5000$ points. Following prior work, we use the last $20$ shapes as a test set and report the performance on this test set. We compare our results with various LBO based methods~\cite{donati20,deepshells,sharma20} in Table~\ref{table:full} as well as embedding based methods \cite{groueix20183d,ric_linear20} as they are applicable, in principle, to both partial and complete shape matching.

\paragraph{Baselines} We compare with the following two broad approaches that are shown to outperform existing competitors: 

\paragraph{LBO based Methods.} Such baselines \cite{sharma20,deepshells} assume to be given as input a mesh representation of a shape as they rely on LBO. While \cite{donati20,sharma20} directly learn features from raw 3D data similar to our method, they project them into LBO basis. \cite{deepshells} refines pre-computed shot descriptors\cite{shot} to learn shape matching. We provide results after refining the point to point map with ZoomOut\cite{melzi2019zoomout} where applicable for all the methods. Note that in presence of outliers and noise, such a refinement makes the resulting point to point map worse and thus, for FAUST-N, we do not apply it. \cite{deepshells} already has a refinement built in their architecture.

\paragraph{Embedding based Methods.} 3D-Coded\cite{groueix20183d} and Marin et al.\cite{ric_linear20} are considered state-of-the-art in learning correspondence directly from point cloud representation without relying on LBO. Note that the baseline~\cite{ric_linear20} is somewhat different from others since it requires and thus, learns both basis functions and probe functions (feature descriptors). 

\begin{figure}[h]
  \minipage{0.27\textwidth}
  \centering
  \includegraphics[width=\linewidth]{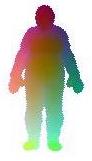}
  \subcaption{Source shape}
\endminipage
\minipage{0.27\textwidth}
 \centering
  \includegraphics[width=\linewidth]{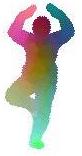}
  \subcaption{\cite{ric_linear20}}
\endminipage
\minipage{0.27\textwidth}%
\centering
 \includegraphics[width=\linewidth]{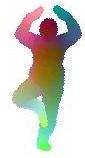}
  \subcaption{Ours.}
\endminipage
\caption{On the left, we show the source shape. In the middle, we transfer a color function on a target shape using Marin et al. \cite{ric_linear20} whereas on the right, we show the transfer using our results.}
\label{fig:match1}
\end{figure}

\paragraph{Ours.} For all results in this paper, we denote our method with symmetry supervision as Ours-sym-Sup and without symmetry supervision as Ours-sym-Unsup. Here symmetry supervision means the access to the ground truth self-symmetry map that is publicly available for Faust-R point clouds. While our method already achieves good performance without symmetry ground truth during training, we include Ours-sym-Sup to show the additional gain brought in by additional symmetry supervision during training.
\begin{table}
\caption{Avg. Geodesic Error for Shape Matching on FAUST}
\begin{center}
\begin{tabular}{l c c c}
\hline
Method $\backslash$ Dataset & FAUST-R & FAUST-N\\
\hline\hline
GeomFM\cite{donati20}+Zo & \textbf{19} &320\\
DeepShell\cite{deepshells} & \textbf{17} & 240\\
Sharma-Ovsjanikov\cite{sharma20}+Zo & 50 & 280\\
\hline
3D-Coded\cite{groueix20183d} & 25 & \textbf{68}\\
Marin et al.\cite{ric_linear20}  & 70&  90 \\
Marin et al.\cite{ric_linear20} +Zo & 50& - \\
Ours-sym-Sup. & 33  &\textbf{58}\\
Ours-sym-Sup.+Zo & \textbf{18} &-\\
Ours-sym-Unsup. & 50  & \textbf{69}\\
Ours-sym-Unsup.+Zo & \textbf{18} & -\\
\hline
\end{tabular}
\end{center}
\label{table:full}
\end{table}
\vspace{-.1cm}
\paragraph{Results and Discussion.} As evident in Table~\ref{table:full}, we obtain competitive performance on FAUST-R. LBO  eigen functions already form a good basis for shapes and thus, prior work based on it obtains impressive performance. However, performance of this line of work degrades significantly under noise, as shown in the Table~\ref{table:full} and also in \cite{ric_linear20}. Thus, our method is significantly more resilient to noise than LBO-based methods. Compared to embedding-based approaches, we obtain slightly better accuracy. In particular, our symmetry-unsupervised version, Ours-sym-Unsup, obtains slightly better performance than our main baseline \cite{ric_linear20}. We also provide a qualitative example to show comparison with \cite{ric_linear20} in Figure \ref{fig:match1}. Note that the right foot is mismatched in Marin et al. whereas we transfer it comparatively well without left-right ambiguity. We note that 3D-Coded is also resilient to noise in point clouds and achieves competitive performance in both scenarios.

We also note that \cite{sharma20} proposes to align the shapes rigidly and shows that manual rigid alignment resolves some symmetry problems arising in shape matching. In this work, we go further and provide an alternative based on a principled approach. 
\vspace{-.2cm}
\begin{table}
\caption{Avg. Geodesic Error on partial SHREC benchmarks}
\begin{center}
\begin{tabular}{l c c}
\hline
Method $\backslash$ Dataset &  Holes & Cuts  \\
\hline\hline
Litany et al.\cite{litany17} & 16  &  13\\
Sharma-Ovsjanikov \cite{sharma20} & 14  &   16\\
Marin et al.\cite{ric_linear20}& 12  & 15 \\
Ours-sym-UnSup. & \textbf{10}  &   \textbf{12}\\
\hline
\end{tabular}
\end{center}
\label{table:partial}
\end{table}
\vspace{-.1cm}
\paragraph{Partial Shape Matching.} For a fair comparison with \cite{sharma20,litany17}, we follow the same experimental setup and test our method on the challenging SHREC’16 Partial Correspondence dataset \cite{cosmo2016matching}. The dataset is composed of
200 partial shapes, each containing about few hundreds to 9000 vertices, belonging to 8 different classes (humans and animals), undergoing nearly-isometric deformations in addition to having missing parts of various forms and sizes. Each class comes with a “null” shape in a standard pose which is used as the full template to which partial shapes are to be matched. The dataset is split into two sets, namely cuts (removal of a few large parts) and holes (removal of many small parts). We use the same test set following \cite{sharma20}. Overall, this test set contains $20$ shapes each for cuts and holes datasets chosen randomly from the two sets respectively. In addition to \cite{ric_linear20}, we compare with the following two baselines: 

\noindent \textbf{Sharma \& Ovsjanikov}\cite{sharma20}. This baseline relies on learning LBO alignment and thus, is dependent on class and needs to be retrained for each of the 8 classes. We include their results even though our results are class agnostic and thus, significantly more robust and efficient. We obtain these results by running the code provided by the authors.

\noindent \textbf{Litany et al.}\cite{litany17}. This baseline is not learning based and relies on hand crafted features and an expensive optimization scheme on the Stiefel manifold for every pair of shapes at test time. Thus, in terms of computation and ground truth map requirement, it is most expensive.

\paragraph{Results and Discussion}
\label{subsec:quant_res}
\vspace{-.2cm}
\begin{wrapfigure}{r}{0.45\textwidth}
\centering
\minipage{0.22\textwidth}
  \includegraphics[width=\linewidth]{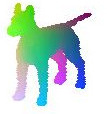}
\endminipage
\minipage{0.22\textwidth}%
  \includegraphics[width=\linewidth]{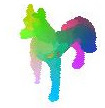}
\endminipage
\caption{Left shows a source shape and the right shows the correspondence map color coded.}
\label{fig:part1}
\end{wrapfigure}
\vspace{-.1cm}
We present our findings on partial shape matching in Table \ref{table:partial} where we obtain superior performance on both benchmark datasets for partial shape matching. We would like to stress that baseline such as \cite{sharma20,litany17} are class specific and need to be trained each time for a class whereas our method is class agnostic and can obtain good results with a fraction of computational time. Similarly, \cite{ric_linear20} trains a similar network as ours two times. First, it learns an embedding with a network similar to ours, followed by a similar network training to compute the optimal linear transformation between the two embeddings. Moreover, the test phase also requires running the network twice. Therefore, our method is at least twice faster than this baseline in computational complexity. We provide a qualitative result of our method in Figure \ref{fig:part1}.

\subsection{Ablation Study}\label{subsec:ablation} 
In Table \ref{table:ablation}, we ablate the overall performance and validate our claim on two different correspondence map coupling via the commutativity loss.
\begin{table}
\caption{Ablation Study for Shape Matching}
\begin{center}
\begin{tabular}{l c}
\hline
Method $\backslash$ Dataset &  FAUST-R  \\
\hline\hline
$NN$  & 61  \\
$NN + NN_{sym}$ & 108  \\
$NN +$ comm.(sup) & 33 \\
$NN +$ comm.(unsup) & 50 \\
\hline
\end{tabular}
\end{center}
\label{table:ablation}
\end{table}  

\textbf{NN}: This baseline ablates the overall performance of our method and quantifies the gain brought in by the pairwise point to point ground truth map alone during training. It shows the performance if we learn an embedding by just projecting the shapes into a canonical space using point to point pairwise map.

\textbf{NN+$NN_{sym}$}: This baseline shows the results obtained for shape matching with strong supervision i.e. instead of using commutativity loss defined in Eq \ref{eq:comm}, we replace it with a  nearest neighbour loss that preserves the ground truth symmetry map for each shape. This baseline is most important to quantify the coupling effect brought in by our commutative loss.
\begin{wrapfigure}{r}{.5\textwidth}
\centering
\minipage{0.24\textwidth}
  \includegraphics[width=\linewidth]{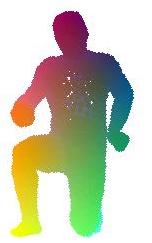}
\endminipage
\minipage{0.23\textwidth}%
  \includegraphics[width=\linewidth]{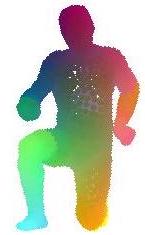}
\endminipage
\caption{Left shows a source shape and the right shows our self-symmetry map.}
\label{fig:sym1}
\end{wrapfigure}
\textbf{NN + comm.(sup)}: This baseline combines the above baseline with the commutativity loss with symmetry supervision and quantifies the gain brought in by commutativity loss (Eq \ref{eq:comm}) in supervised setting. 

\textbf{NN + comm.(unsup)}: shows the gain brought in by coupling with a commutativity loss in an unsupervised way (Eq \ref{eq:comm_uns}) and represents Ours-sym-Unsup.

\paragraph{Discussion} Our ablation study shows the individual importance of the two loss functions. We note that the performance gains brought in by commutative loss on self-symmetry embeddings are significant. More specifically, as evident in Table \ref{table:ablation}, using just the nearest neighbour loss on a self-symmetry map and a pairwise map, denoted as NN + $NN_{sym}$ in Table \ref{table:ablation}, overfits badly as there is no explicit information transfer or constraint between the two maps. 
 \begin{table}
\caption{Avg. Geodesic Error for self-symmetry maps}
\begin{center}
\begin{tabular}{l c c c c}
\hline
 \small{Method $\backslash$ Dataset} & \small{Faust-A}& \small{Scape-A}& \small{Faust-N} & \small{Scape-N}\\
\hline\hline
 \small{Nagar-Raman}\cite{Nagar18} & 34  & 60 & -&-\\
\small{Ren et al.+Zo}\cite{jing_sig20} & \textbf{19} & \textbf{54}& 166& 193\\
 \small{Our-sym-Sup.+Zo} & \textbf{29}  & \textbf{63}& 58& 88\\
 \small{Our-sym-Unsup+Zo} & 50 & 75 & 66&95\\
\hline
\end{tabular}
\end{center}
\label{table:sym}
\end{table}
\subsection{Symmetry Detection}\label{subsec:symmetry} This subsection evaluates our method on the task of symmetry detection in non-rigid shapes. We evaluate it on FAUST aligned dataset (FAUST-A), SCAPE-A as well as its noisy version. We use the usual train-test split where we test on the last $20$ shapes for FAUST-A and last $12$ shapes for SCAPE-A. We show the comparative results in Table \ref{table:sym} where we compare with multiple baselines.

\begin{wrapfigure}{r}{.48\textwidth}
\centering
\minipage{0.23\textwidth}
  \includegraphics[width=\linewidth]{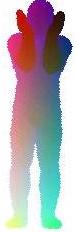}
\endminipage
\minipage{0.23\textwidth}%
  \includegraphics[width=\linewidth]{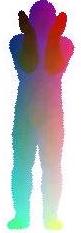}
\endminipage
\caption{Left shows a source shape and the right shows our self-symmetry map.}
\label{fig:sym2}
\end{wrapfigure}
\vspace{-.1cm}
In particular, Ren et al.~\cite{jing_sig20} is considered state-of-the-art and heavily relies on LBO to estimate self-symmetry maps. We show our results with both symmetry supervision, denoted as Ours-sym-Sup as well as without symmetry supervision denoted as Ours-sym-Unsup in Table \ref{table:sym}. Similar to Ren et al.\cite{jing_sig20}, we also refine our point to point map by applying zoomout to initial maps. For the noisy setting, we simply show results as such and do not apply zoomout refinement as it is based on LBO that is unreliable in a noisy setup. We provide a qualitative example from SCAPE-A in Figure \ref{fig:sym1} and from FAUST-A in Figure \ref{fig:sym2} to illustrate our results. 

\paragraph{Discussion} Table \ref{table:sym} shows that axiomatic approach of Ren et al.\cite{jing_sig20} obtains slightly better performance than us on both FAUST-A and SCAPE-A. However, in the presence of noise, its performance suffers significantly. We also remark that we are not aware of any other work that investigates the performance of axiomatic approach for symmetry detection in the presence of outliers. Our method also undergoes a decrease in accuracy. However, our approach is still resilient to noise and performs significantly better than Ren et al.

\begin{figure}[h]
   \centering
\includegraphics[width=.99\linewidth]{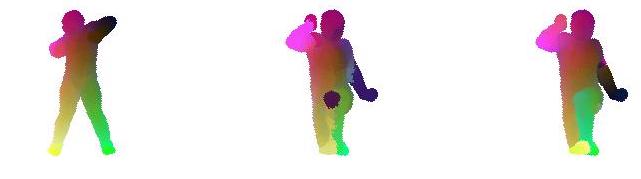}
\caption{Failure Case: On the left, we show the source shape. In the middle, we transfer a color function on a target shape using Marin et al. \cite{ric_linear20} whereas on the right, we show our result.}
\label{fig:failure}
\end{figure}

\paragraph{Failure Case} We show a failure case from SCAPE-A in Figure \ref{fig:failure} where our method finds it challenging to disambiguate symmetry. It maps the right foot of source shape to the left foot on target shape. Marin et al. still performs worse than us as it fails to disambiguate the  lower leg of source shape from the  lower right leg of target shape. Human poses are quite diverse and this example shows a failure for symmetry detection when training with small data.

\vspace{-.1cm}
\section{Conclusion} In shape correspondence literature, partial shape matching and full shape matching are generally tackled by two different sets of methods which obtain impressive results in one of the two respective domains. Similarly, symmetry detection and shape matching are also learned or modelled separately. We presented a simple, general but promising approach that provides a unifying framework and reduces pairwise as well as self-symmetry map estimation to a nearest neighbour search in a canonical embedding. Our approach is significantly more resilient to noise than methods based on predefined basis/embedding functions. We believe our key idea of coupling a self-symmetry and a pairwise map via commutativity will encourage future work to explore similar constraints in unsupervised or weakly supervised learning of canonical embeddings. 


\bibliographystyle{splncs04}
\bibliography{main}
\end{document}